\documentclass[10pt,twocolumn,letterpaper]{article}

\usepackage{cvpr}              
\usepackage{graphicx}
\usepackage{amsmath}
\usepackage{amssymb}
\usepackage{booktabs}
\usepackage{algorithm}
\usepackage{algpseudocode}
\usepackage{array}
\newcolumntype{P}[1]{>{\centering\arraybackslash}p{#1}}
\newcolumntype{M}[1]{>{\centering\arraybackslash}m{#1}}
\usepackage[accsupp]{axessibility}  
\usepackage{nopageno} 
\usepackage[pagebackref,breaklinks,colorlinks]{hyperref}

\usepackage[capitalize]{cleveref}
\Crefname{section}{Sec.}{Secs.}
\Crefname{section}{Section}{Sections}
\Crefname{table}{Table}{Tables}
\Crefname{table}{Tab.}{Tabs.}

\def\cvprPaperID{3} 
\def\confName{CVPRW}
\def\confYear{2022}

\begin{document}

\title{Facial Expression Recognition using Squeeze and Excitation-powered Swin Transformers}

\author{Arpita Vats\\
Santa Clara University\\
{\tt\small avats@scu.edu}
\and
Aman Chadha\\
Amazon, Stanford University \\
{\tt\small hi@aman.ai}
}
\maketitle

\begin{abstract}
The ability to recognize and interpret facial emotions is a critical component of human communication, as it allows individuals to understand and respond to emotions conveyed through facial expressions and vocal tones. The recognition of facial emotions is a complex cognitive process that involves the integration of visual and auditory information, as well as prior knowledge and social cues. It plays a crucial role in social interaction, affective processing, and empathy, and is an important aspect of many real-world applications, including human-computer interaction, virtual assistants, and mental health diagnosis and treatment. The development of accurate and efficient models for facial emotion recognition is therefore of great importance and has the potential to have a significant impact on various fields of study.The field of Facial Emotion Recognition (FER) is of great significance in the areas of computer vision and artificial intelligence, with vast commercial and academic potential in fields such as security, advertising, and entertainment. We propose a FER framework that employs Swin Vision Transformers (SwinT) and squeeze and excitation block (SE) to address vision tasks. The approach uses a transformer model with an attention mechanism, SE, and SAM to improve the efficiency of the model, as transformers often require a large amount of data. Our focus was to create an efficient FER model based on SwinT architecture that can recognize facial emotions using minimal data. We trained our model on a hybrid dataset and evaluated its performance on the AffectNet dataset, achieving an F1-score of 0.5420, which surpassed the winner of the Affective Behavior Analysis in the Wild (ABAW) Competition held at the European Conference on Computer Vision (ECCV) 2022~\cite{Kollias}.

\vspace{5pt}

\end{abstract}
\section{Introduction}
\label{sec:intro}
Facial Emotion Recognition (FER) is one of the major areas of research. (FER) is a field of study in computer vision and artificial intelligence that focuses on the detection and interpretation of emotions expressed through facial expressions. FER technology uses computer algorithms to analyze images or videos of faces and identify emotions such as happiness, sadness, anger, fear, surprise, and disgust.
FER has the potential to impact a wide range of applications, including psychology and neuroscience research, marketing, human-computer interaction, and security. In psychology and neuroscience, FER can help researchers better understand the emotions that drive human behavior. In marketing, FER can be used to gauge consumer emotions and preferences. In human-computer interaction, FER can be used to create more natural and intuitive interfaces that respond to human emotions. In security, FER can be used for authentication, surveillance, and emotional profiling. Faces analysis indicates recognizing the angle and expression of a human being independently of the immersive environment it could be, and ambiguous emotions are the cornerstone of the problem. Understanding human emotion also plays a vital role in emotional intelligence. Facial expression is a primal, impactful, and ubiquitous means by which humans communicate their feelings and motives. The intricate and nuanced movements of facial muscles, even the most subtle changes, can convey a range of emotions that are universally understood and instinctively recognized by people from all cultures and backgrounds. From joy and sadness to anger and fear, facial expressions serve as a fundamental tool for human interaction and are a crucial component of nonverbal communication. \cite{darwin_2013, Tian}. We seek to analyze how the Swin transformer (Swin-T) performs on this task, comparing our model with the state-of-art models on hybrid datasets, taking into account the lack of inductive bias proper for Vision Transformer (ViT). ViT is a transformer-based architecture for computer vision tasks such as image classification, segmentation, and object detection. The paper ~\cite{dosovitskiy2021an} demonstrates that  ViT outperforms existing state-of-the-art models on several benchmark datasets, and provides insights into how the architecture works and why it is effective. It uses self-attention mechanisms to dynamically attend to essential regions in an image, allowing them to capture complex relationships between objects. Using transformers for image recognition makes it possible to achieve strong results on image recognition tasks while using less memory and computational resources than traditional CNN. We offer an overview of the following aspects: 
\begin{itemize}
    \item Data composition: Understanding the data composition of different datasets with high data variables, and merging them into a unique dataset.
    \item Data integration: Integrating data from various sources to create a unified dataset.
    \item Data analysis: Analyzing the features of each subset of data, including some attributes and metadata to change for normalized samples.
    \item Data preprocessing: Preparing the data for manipulation and augmentation, including techniques such as normalization, scaling, and augmentation.
    \item Dataset split: Splitting the dataset into three subsets with some common features, such as image format, size, and the number of channels.
    \item Face detection and cropping: Configuring models for face detection and cropping procedures.
    \item Model evaluation: Assessing the outcomes of models through the utilization of diverse performance metrics, including accuracy, precision, recall, and F1-score, is a critical aspect of evaluating their effectiveness.
    \item Results analysis: Analyzing the results of the models to understand the strengths and weaknesses of the transformers for Facial Emotion Recognition.
\end{itemize} 
In this work, we presented a Facial Emotion Recognition (FER) framework in this work. Our approach is based on SwinT and squeeze and excitation block (SE). To develop an efficient FER model with the ability to detect facial emotions using a small amount of data, we utilized a transformer model with an attention mechanism and a sharpness-aware minimizer (SAM). Additionally, we made a unique contribution by using a hybrid dataset for training and evaluating the model's performance on the AffectNet dataset, achieving an F1-score of 0.5420. The effectiveness of our approach was demonstrated by outperforming the winner of the Affective Behavior Analysis in-the-wild (ABAW) Competition held in conjunction with the European Conference on Computer Vision (ECCV) 2022.
\footnote{Work does not relate to the position at Amazon}

\begin{table}[tb]
\centering
\caption{ActiViTy Classes.}\label{tbl:actiViTies}
\footnotesize
\begin{tabular}{llll} 
\toprule
\textbf{ID} & \textbf{Class} \\ 
\toprule
0  & Fear\\
1  & Sadness\\
2  & Happy \\
3  & Anger  \\
4  & Disgust\\
5  & Surprise \\
6 & Neutral\\
\bottomrule
\end{tabular}
\end{table}

\section{Related Works}
\label{sec:related}
Deng~\etal~\cite{Deng} suggested a technique for multi-task learning in the presence of missing labels. To balance the dataset, they proposed a method that utilized the ground truth labels of all three tasks to train a teacher model and then used the output of the teacher model as soft labels for the student model. They used both the soft labels and the ground truth labels to train the student model.

Kuhnke and Rumberg~\etal~\cite{Kuhnke_2020} proposed a two-stream model that incorporated audio and image streams. They fed these streams separately into a CNN network, then utilized temporal convolutions on the image stream. Additionally, they utilized facial alignment and correlations between different emotional representations to improve their model's performance.

Thinh~\etal~\cite{Thinh} introduced a deep learning model that used ResNet50 \cite{He} as its backbone, with pre-trained weights from ImageNet~\cite{Deng}. They employed VGGFace2 for emotion recognition, aiming to speed up and enhance the training process.

Zhang~\etal~\cite{article} proposed a method for multi-task emotion recognition that takes into account the intrinsic association between the different emotional representations. They noted that despite the different psychological philosophies behind these representations, there is evidence that they are linked to each other. For example, similar facial muscle movements (action units) tend to indicate similar emotions, and most previous works on multi-task emotion recognition have ignored this fact by modeling different tasks in parallel branches. The proposed method instead uses a streaming structure to model the recognition process serially, going from local action units to global emotion states, and adjusting the hierarchical distributions on different feature levels. This approach is designed to better capture the interdependent relationships between the different emotional representations.

DAN, a facial recognition model introduced by Wen~\etal~\cite{wen}, comprises three key components: Feature Clustering Network (FCN), Multi-head cross Attention Network (MAN), and Attention Fusion Network (AFN). FCN is responsible for feature extraction using a large-margin learning approach to maximize class separability. MAN, on the other hand, utilizes several attention heads to attend to multiple facial areas simultaneously, building an attention map for these regions. Finally, AFN combines the attention maps by distracting attention to multiple locations before fusing them into a comprehensive map.

The current state-of-the-art approach for emotion recognition using the AffectNet dataset was proposed by Andrey~\etal~\cite{Andrey}. Their method involves applying face detection, tracking, and clustering techniques to extract face sequences from each frame. Subsequently, a single neural network is used to extract emotional features from each frame.
\begin{figure*}[h]
  \centering
  \includegraphics[width=\linewidth]{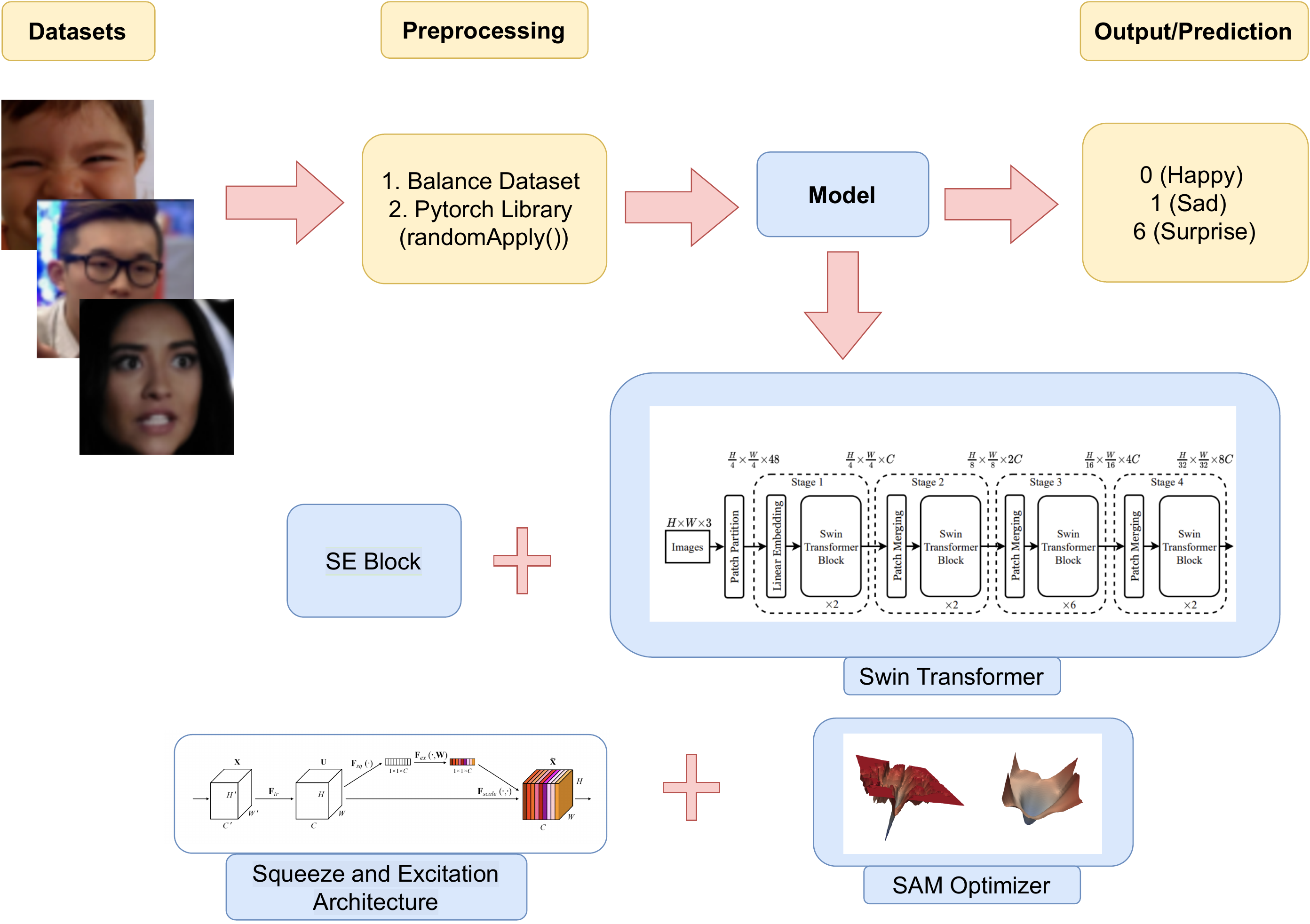}
  \caption{Facial Emotion Detection using \texttt{SwinT} with SE Block.}
  \label{fig:pose}
\end{figure*}
\section{Methodology}
\label{sec:methods}

\Cref{fig:pose} Depicts our framework. The proposed architecture for facial emotion recognition utilizes a SwinT model augmented with a SE layer before the Swin Transformer. Swin Transformer, also known as SwinT, is a hierarchical Transformer that computes image representation using Shifted windows, allowing for cross-window connection and efficient self-attention computation. This hierarchical approach permits modeling at various scales while maintaining linear computational complexity with respect to image size. The model's primary task is to predict basic facial emotions, and the inclusion of the SE layer enhances its robustness, maximizing intra-distance between clusters. The shifted windowing scheme increases computational efficiency by limiting self-attention computation to non-overlapping local windows, resulting in greater speed and scalability for large datasets.

\subsection{Swin Transformer}
\label{sec:methods:driver}

In recent years, the Transformer architecture has gained widespread recognition and adoption in the field of machine learning, especially in Natural Language Processing (NLP). Introduced in 2017, the Transformer architecture has revolutionized the way in which sequence-to-sequence tasks are performed. Prior to this, recurrent neural networks (RNNs) were commonly used for sequence tasks, but the Transformer architecture offered a more efficient and parallelizable solution.

Vision Transformer(ViT), a variant of the Transformer architecture that emerged in 2020, has revolutionized computer vision tasks, particularly image recognition, by offering a fresh approach to conventional models that utilize convolutional neural networks (CNNs). ViT leverages self-attention mechanisms instead of convolutions, enabling the network to dynamically focus on critical regions within an image. Impressively, ViT has achieved remarkable outcomes in various image recognition tasks, surpassing traditional CNNs while utilizing fewer memory and computational resources.

By harnessing vision transformers as presented in (2021)~\etal~\cite{Liu}(2021), we have successfully classified eight human emotions, including anger, contempt, disgust, fear, happiness, neutral, sadness, and surprise, by fine-tuning a pre-trained ImageNet model. The attention mechanism plays a crucial role in this model by extracting vital features from the input through a standard query, key, and value structure. The similarity between queries and keys is established through matrix multiplication, followed by the application of the softmax function to the outcome, resulting in the 'attention' mechanism. Our transformer architecture comprises eleven encoders stacked on top of a hybrid patch embedding architecture. Our approach overcomes the lack of an inductive bias problem, which is a concern in Vision Transformers as they have significantly fewer image-specific inductive biases than CNNs.
Swin Transformer (SwinT), an avant-garde architecture tailored for computer vision, represents a novel amalgamation of two robust models, Convolutional Neural Networks (CNNs) and Transformers, combining their unparalleled strengths to create a powerhouse of a model. SwinT surpasses the previous state-of-the-art Vision Transformers (ViT) by introducing a multi-scale approach that effectively captures both local and global features with exceptional precision, making it an unparalleled choice for complex computer vision tasks that necessitate a fine-grained and global understanding of visual data. SwinT's innovative integration of these models not only facilitates efficient and accurate image recognition but also significantly reduces computational costs.

The SwinT architecture is a combination of convolutional and self-attention mechanisms, with a unique switching mechanism that enables it to capture fine-grained details and high-level semantic information in images. Its impressive performance in various computer vision benchmarks suggests its potential for driving further progress in the field.

To construct a SwinT block, the multi-head self-attention (MSA) module in a Transformer block is replaced with a shifted window-based MSA module. This module is followed by a 2-layer MLP with GELU nonlinearity in between. Each MSA module and MLP is preceded by a LayerNorm (LN) layer, and a residual connection is applied after each module. The input RGB image is divided into non-overlapping patches using a patch-splitting module, and each patch is treated as a token. The feature of each patch is constructed by concatenating the raw pixel RGB values, resulting in a patch dimension of 48, accounting for the 3 channels of a 4 $\times$ 4 patch. A linear embedding layer is applied to this raw-valued feature to project it to an arbitrary dimension C. This hierarchical architecture has linear computational complexity with respect to image size and is flexible enough to model at various scales.

The SwinT block's unique combination of convolutions and self-attention mechanisms has shown great promise in various computer vision tasks, such as image classification and object detection. With its shifted window-based MSA module, it has the potential to outperform other state-of-the-art models in the field.

\subsection{Datasets}
\label{dataset}
During the development of our robust model, we encountered a significant obstacle - the scarcity of sufficient data. Many datasets, often accessible only for research purposes, remain out of reach for students, limiting the amount of data we could use. We resorted to using open-source data platforms such as Kaggle, which provided some samples but fell short of the amount required for effective training of transformers. To address this challenge, we devised a plan to augment the limited data using various techniques, thereby increasing the size of the final datasets.

Our approach involved utilizing several datasets, each with its unique set of characteristics and limitations. The FER-2013 dataset, for example, contained around 40,000 facial RGB images with varying expressions, restricted to a size of 48 $\times$ 48. The dataset labels were classified into seven primary types, including Fear, Sadness, Happy, Anger, Disgust, Surprise, and Neutral. We observed a significant imbalance in the data across the different expression categories, with Disgust expression having only 600 samples, whereas the remaining labels had almost 5,000 samples each.

Another dataset we used was the CK+ dataset, an extended version of the Cohn-Kanade dataset. It contained images from 593 video sequences of 123 different subjects with diverse genders and heritages, ranging from 18 to 50 years old. Each video sequence depicted a facial transition from a neutral expression to a specific peak expression, recorded at 30 frames per second (FPS) and with resolutions of either 640 $\times$ 490 or 640 $\times$ 480 pixels. However, we could only access a portion of the complete dataset, containing 1000 images with high variability obtained from a Kaggle repository, highlighting the challenge of limited data availability.

Lastly, we also used the AffectNet dataset, which consists of an extensive collection of 60,000 facial expression images classified into eight different classes, including neutral, happy, angry, sad, fear, surprise, disgust, and contempt. The dataset also includes intensity measures of valence and arousal associated with each expression, adding an extra layer of complexity and providing additional information for the model to learn from.

In conclusion, despite the limited availability of data, we employed various techniques to augment the samples we had, utilizing multiple datasets with their unique characteristics and limitations. This enabled us to increase the size of the final datasets and train a robust model that could accurately classify facial expressions.

With each dataset focusing on RGB channels for coloring and having different sizes and image extensions, the overall data size amounts to approximately 2 GB. To handle this diverse dataset, it is imperative to establish a standard format that allows for efficient management of the data. Consequently, we implemented various fine-tuning techniques, as described in the preprocessing section, to manage the data effectively and ensure the optimal performance of the model.


\begin{figure}[h]
  \centering
  \begin{tabular}{cc}
   \hspace{-10pt} \includegraphics[width=\linewidth]{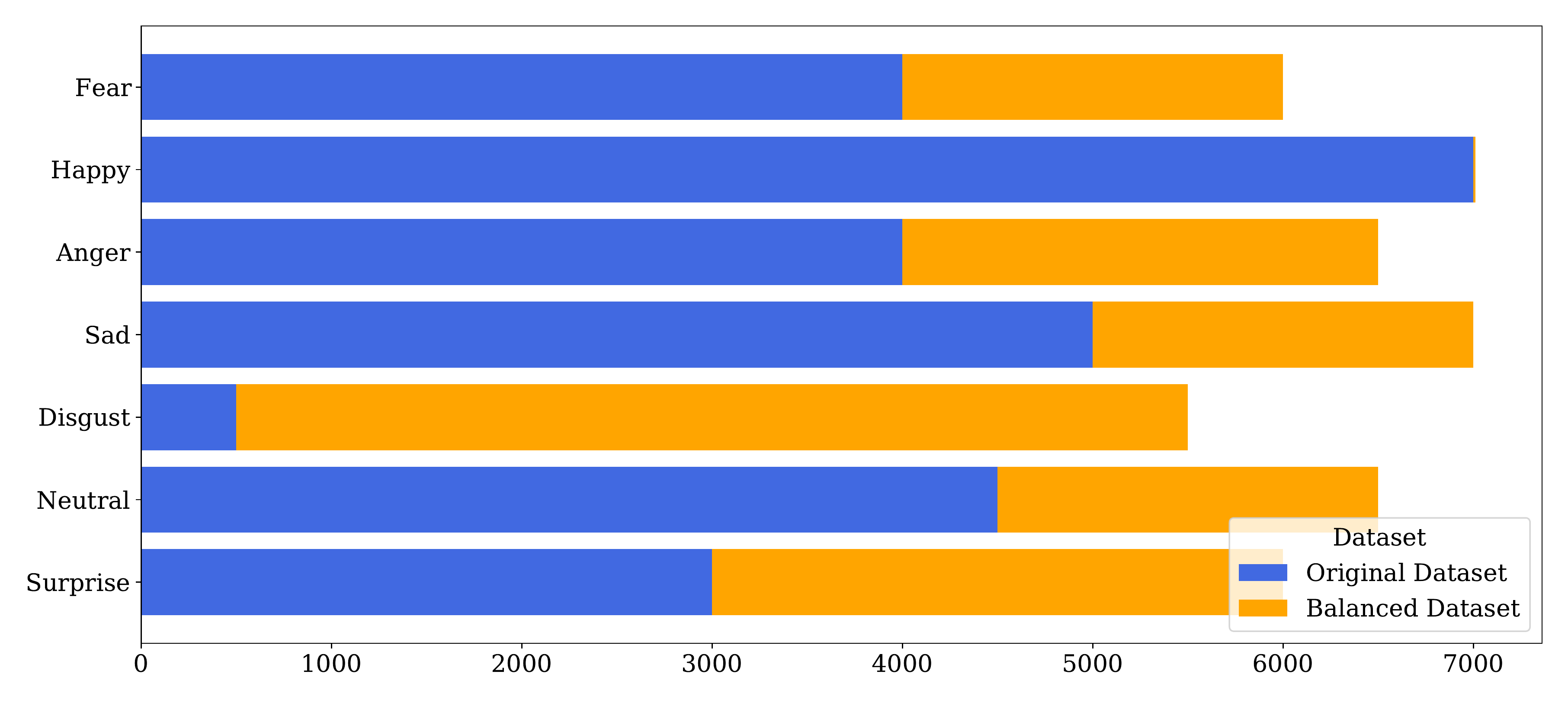}
  \end{tabular}
  \caption {Class-level sub-population statistics for the final dataset after balancing.}
  \label{fig:datasets}
\end{figure}
\subsection{Preprocessing}
\label{sec:methods: features}
During the data collection phase of our project, we sourced data from multiple sources and amalgamated them to create a dataset. However, we found that the dataset was imbalanced as the training set had insufficient samples for each class, while the validation and testing sets had equal samples for every category. To overcome this challenge, we utilized data augmentation methods to increase the number of samples for each category and removed any excessively generated images. This approach led to the creation of a final dataset with an equal number of samples for each class, thereby achieving balance across the entire dataset.

Despite our efforts to include open-source data, we encountered a challenge with the \texttt{contempt} and \texttt{disgust} classes, which had limited amounts of data. To overcome this issue, we leveraged data augmentation techniques to increase the variance of pixel matrices, effectively expanding the available data. By doing so, we could ensure that the final dataset remained balanced, without the need for oversampling techniques.

In this section, we will describe the data manipulation and merging process of multiple datasets, as well as the various data augmentation techniques used to preprocess the dataset for training. Since we used multiple datasets, we had to integrate them into one with the same dimensions and configuration for the model to use as input. Due to an unbalanced class distribution, we utilized various augmentation techniques, including:

One common technique used to increase the diversity of available data and improve the model's generalization ability is data augmentation. Image rotation, a specific type of augmentation technique, is frequently employed by rotating images by a certain degree, usually ranging from 0 to 360. In our case, we rotated the images up to 10 degrees to standardize the frontal images of FER-2013 and CK+48 datasets to have a similar face orientation to AffectNet faces, without affecting the already rotated images. This technique expands the available data and ensures that the model can recognize and learn facial expressions across various face orientations with high accuracy.

To address the challenge of the Transformer architecture's large data requirements and lack of data, we utilized a variety of augmentation techniques to increase the sample size. We applied several augmentation methods, such as RandomRotation\cite{Boykov} and RandomAutocontrast. These techniques helped the model become familiar with more data, ultimately improving its performance. Additionally, we conducted an ablation experiment that proved the effectiveness of augmentation in improving model performance.\Cref{fig:datasets} shows the unbalanced and balanced dataset used for training after preprocessing and integration of different datasets of facial emotions mentioned in section \cref{dataset} in detail.

\subsection{Model}
\label{sec:methods: classification}


In this section, we presented a Single-Step Detector model used for emotion classification and face cropping, along with its adaptations. First, we resized the images to $224$ $\times$ 224 $\times$ 3 to use them as input for the Transformer model. Finally, we normalized the images with a mean and standard deviation of 0.5 for all channels, as used during SwinT fine-tuning, to recognize eight emotions: anger, contempt, disgust, fear, happiness, neutral, sadness, and surprise. SwinT split the RGB input image into non-overlapping patches, each patch's feature being a concatenation of the raw pixel RGB values with a patch dimension of 4 $\times$ 4 $\times$ 3 = 48. A linear embedding layer projected this feature to an arbitrary dimension C. Swin Transformer blocks as shown in ~\cref{fig:swin Transformer}, multiple Transformer blocks with modified self-attention computation, was applied to these patch tokens, maintaining the number of tokens (H/4 $\times$ W/4) and forming "Stage 1" by combining the linear embedding with the blocks. Patch-merging layers reduced the number of tokens, concatenating features of each group of 2 $\times$ 2 patches, and applying a linear layer to the 4C-dimensional concatenated features. This reduced the tokens by a factor of 4, with the output dimension set to 2C and the resolution maintained at H/8 $\times$ W/8. The same process repeated for "Stage 2", "Stage 3", and "Stage 4", creating a hierarchical representation with resolutions similar to VGGNet and ResNet, allowing the backbone networks' easy replacement for existing vision tasks.

 \begin{figure}[t]
  \centering
  \begin{tabular}{cc}
   \hspace{-1pt} \includegraphics[width=\linewidth]{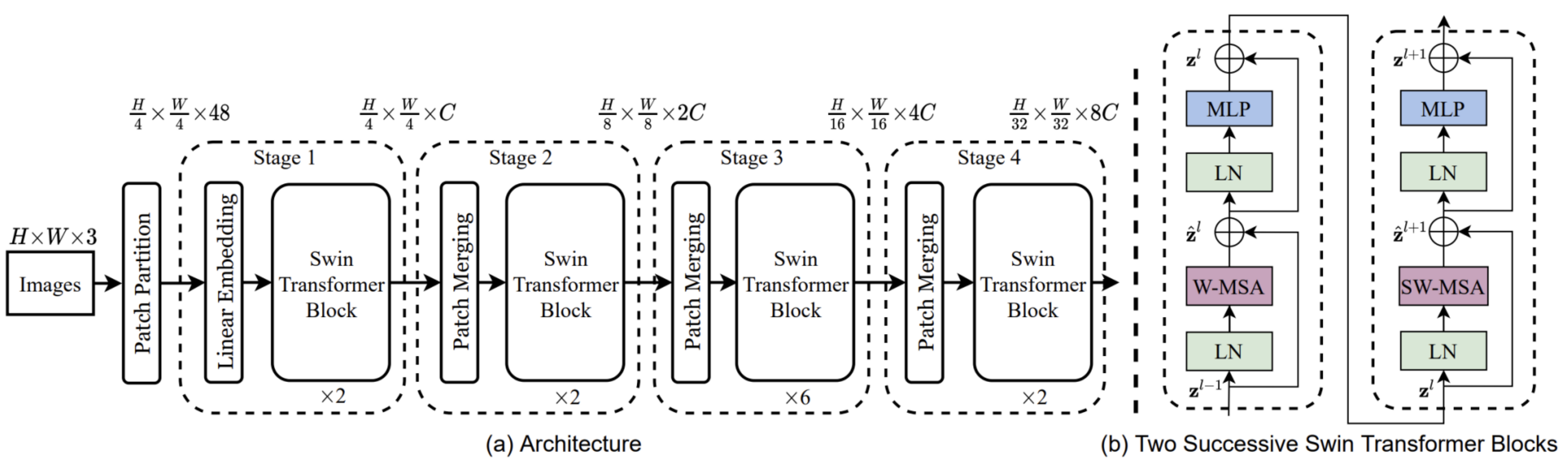}
  \end{tabular}
  \caption{(a) The architecture of a SwinT; (b) two successive SwinT Blocks W-MSA and SW-MSA are multi-head self-attention modules with regular and shifted windowing configurations, respectively \cite{Liu}.}
  \label{fig:swin Transformer}
\end{figure}
\subsection{Squeeze and Excitation}
\label{sec:methods:SE}
The Squeeze and Excitation (SE) block, akin to the self-attention mechanism, is an integral component of attention-based models. However, it comprises fewer parameters than the self-attention block and utilizes only one operation of point-wise multiplication. Initially introduced by~\etal~\cite{shen}(2018) as a channel-wise attention module to optimize CNN architecture, we exclusively use the excitation part of the SE block, as the squeeze part acts as a pooling layer that reduces the dimensionality of 2D-CNN layers, as outlined by~\cite{Aouayeb} (2020).

We apply the SE block atop the Transformer encoder, specifically on the classification token vector. Unlike the self-attention block, which encodes the input sequence and extracts features through the class token within the Transformer encoder, the SE block re-calibrates the feature responses by modeling inter-dependencies among class token channels explicitly. This technique enhances the model's ability to identify and learn significant features by selectively amplifying relevant channels and suppressing irrelevant ones, ultimately leading to improved performance.
\subsection{Transformer with Sharpness-Aware Minimizer}
\label{sec:methods:SAM}
The Sharpness-Aware Minimizer (SAM) algorithm, as proposed by Chen~\etal~\cite{chen}(2020), leverages the intricate geometry of the loss landscape in deep neural networks to enhance their generalization capabilities. Unlike conventional optimization methods that prioritize the individual parameter's loss value, SAM seeks to smooth the loss landscape and minimize both the loss value and curvature simultaneously, resulting in parameters that exhibit uniformly low loss values and linear curvatures on the loss values.

When applied to the Vision Transformer model, SAM can minimize loss values while simultaneously improving training time. Additionally, SAM's optimization function can address the problem of noisy labeling in datasets, a common challenge in datasets like Affectnet. However, it is crucial to note that SAM's effectiveness reduces as the training dataset size increases, which presents a challenge when dealing with unbalanced datasets like Affectnet which have a low number of samples for certain emotions such as contempt and disgust.

Despite the additional computational costs per update, SAM has demonstrated promising results on small datasets and can potentially serve as a valuable tool to enhance the performance of deep neural networks. SAM's ability to optimize the loss landscape's geometry and minimize the impact of noisy labeling in datasets could be instrumental in overcoming some of the challenges associated with training deep neural networks.
\begin{figure}[h]
  \centering
  \begin{tabular}{cc}
   \hspace{-10pt} \includegraphics[width=0.49\linewidth]{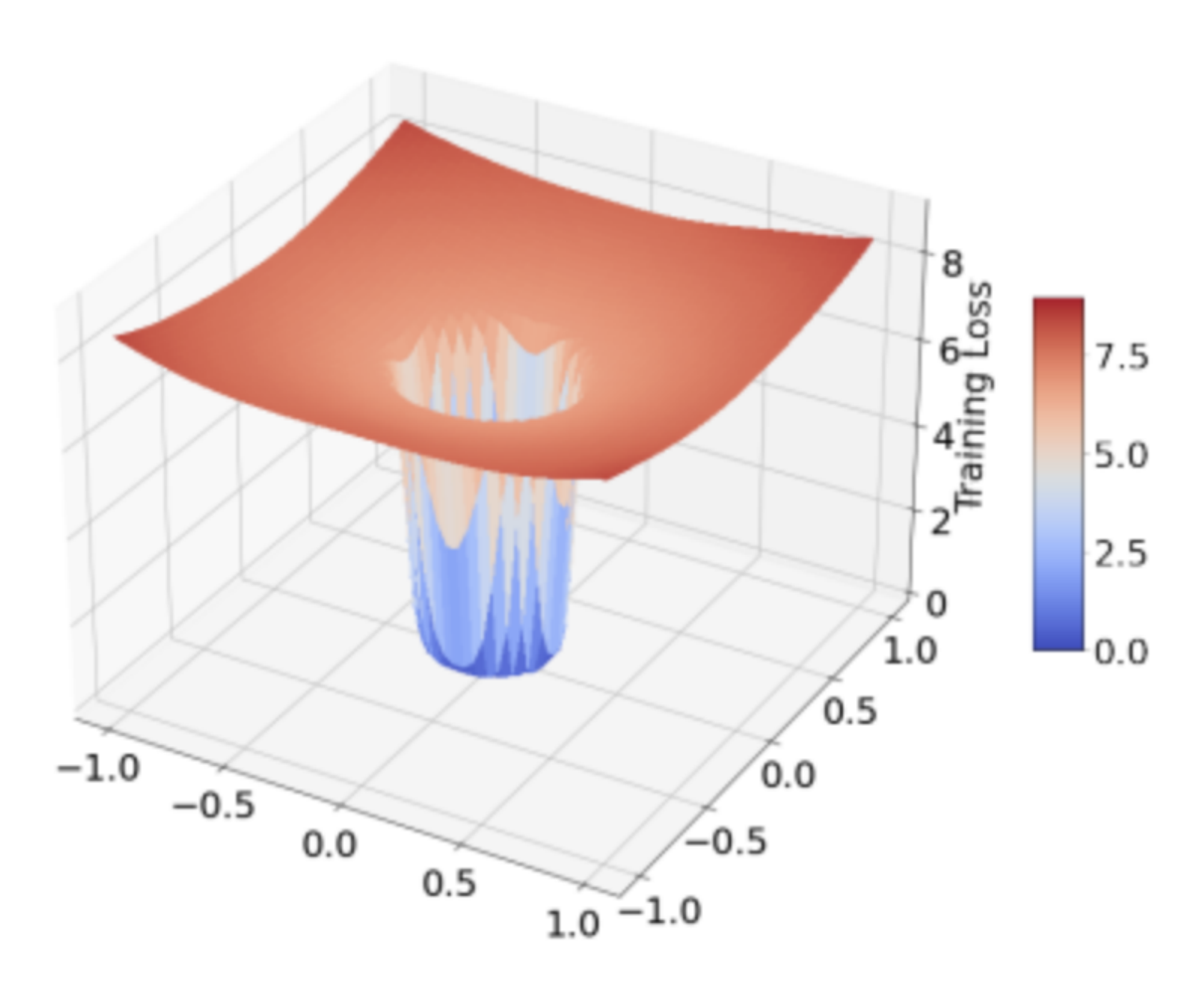}  \\ 
  \includegraphics[width=0.49\linewidth]{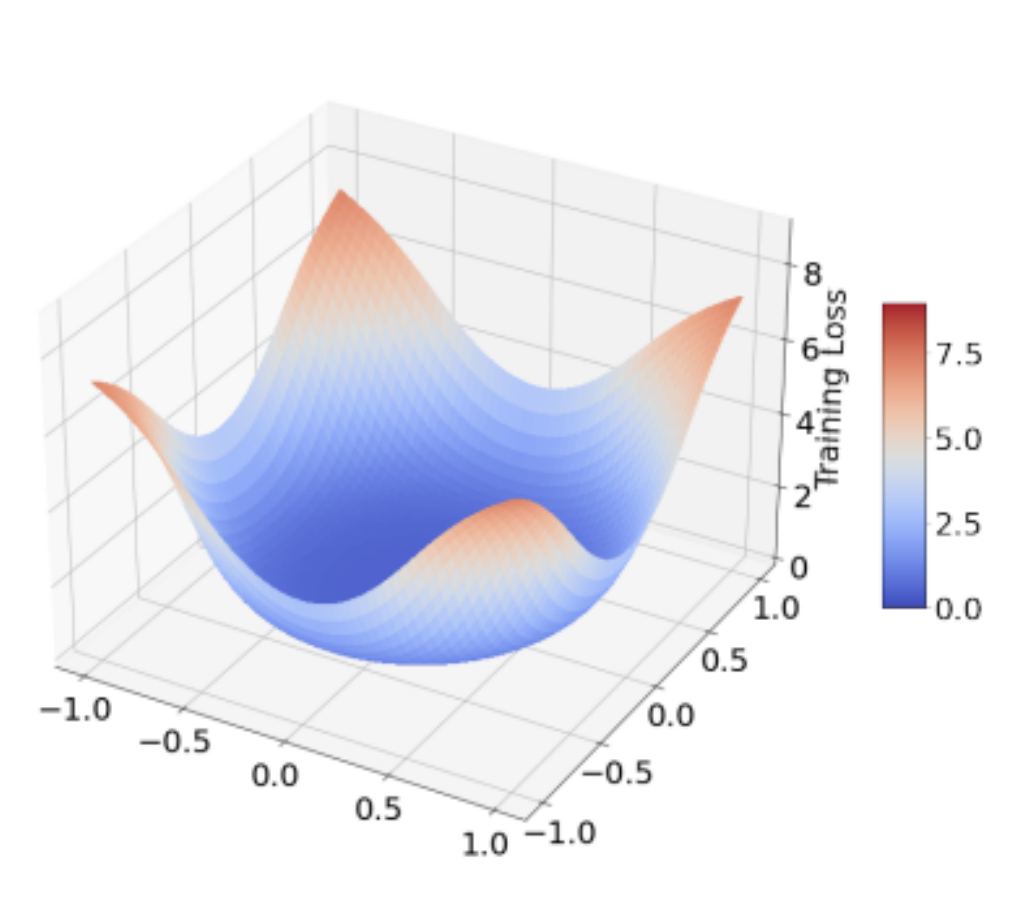} \\ 
  \end{tabular}
  \caption{ Cross-entropy loss landscape on ViT (top) and the same smoothed landscape with the application of SAM (bottom) during the training on ImageNet \cite{chen}.}
  \label{fig:swin Transformer}
\end{figure}

\begin{figure*}[t]
  \centering
  \begin{tabular}{cc} 
  \includegraphics[width=0.49\linewidth]{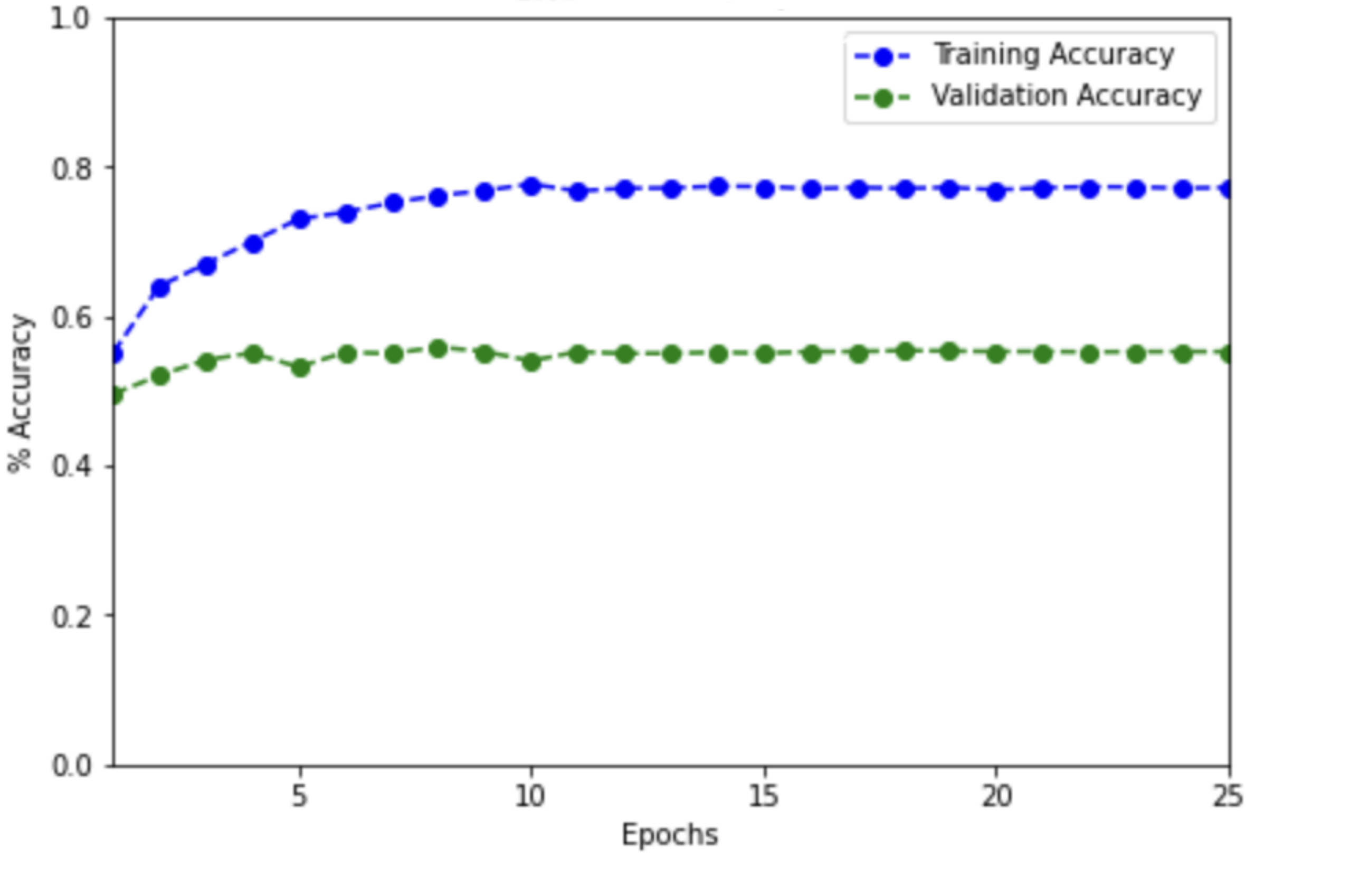} & 
  \includegraphics[width=0.49\linewidth]{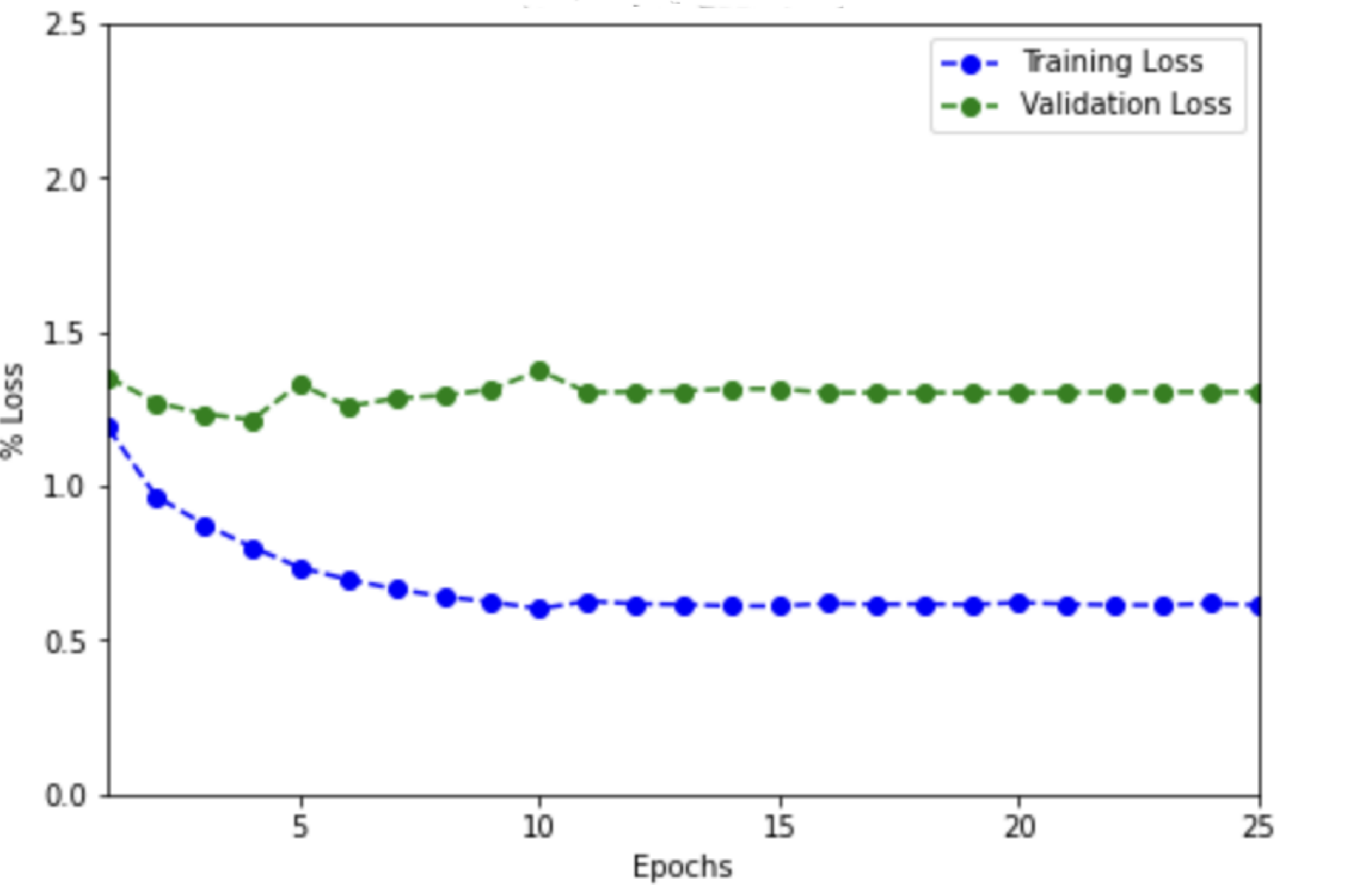} \\
  (a) & (b)
  \end{tabular}
  \caption{(a) Training and Validation Accuracy Swin+SE+SAM (b) Training and Validation Loss  Swin-T+SE+SAM.}
  \label{fig:effectiveness}
\end{figure*}

  
\subsection{Implementation Details}
\label{sec:implementation}
The model we propose leverages a SwinT model pre-trained on ImageNet-1K, with transformer configurations tailored to fine-tuning requirements based on the last layer's dimension. In addition, it provides a randomly weighted version for each structure without any pre-training phase. Our model demonstrates an impressive F1 score of 0.5452. While presenting our experimental evaluation, we will also delve into the potential shortcomings of an alternative method we tested. We executed all our experiments on a system running Ubuntu Linux version 20.04 and equipped with a 12-core Intel(R) Core(TM) i9-7920X CPU @ 2.90GHz, 128 GB RAM, and 4 NVIDIA RTX 3090 24G GPUs. The model implementation is built on PyTorch, utilizing its components as the primary framework. During the preprocessing phase, we meticulously redefined the images' size to conform to 224 $\times$ 224 on three distinct channels (i.e., RGB). Furthermore, we normalized the input data and prepared the samples for the training phase by applying a mean and standard deviation of 0.5 to each channel. The final model weight set is determined by selecting the best validation accuracy from the epochs during the training phase. The fine-tuning phase adapts the model parameters to the FER task using either stochastic gradient descent or sharpness-aware minimizer adaptation, coupled with a cross-entropy loss function. A learning rate scheduler adjusts the initial value for every ten epochs by multiplying it by 0.1, with a momentum of 0.9 applied to increase the training speed and a variable learning rate based on the optimizer chosen in the experiment.

Our experiments were carried out in this environment, with different configurations and SwinT architectures that enable better class separation compared to the CNN baseline architecture. Additionally, the SE block enhances the SwinT model's robustness, as it maximizes the intra-distances between clusters. Interestingly, the features before the SE form more compact clusters with inter-distance lower than the features after the SE, which may suggest that the features before SE are more robust than those after the SE. We tested three different model variants, including \texttt{SwinT}, \texttt{SwinT+SE}, and \texttt{SwinT+SE+SAM}. Based on our empirical observations, we concluded that \texttt{SwinT+SE+SAM} outperforms the other architectures, indicating that our model is capable of accurately recognizing emotions in facial expressions.

\subsection{Results and Evaluation}\label{sec:experiments:Metrics}
We conducted model testing on 4000 diverse samples from AffectNet, using training and validation sets without any data augmentation. In ~\cref{tbl: F1score} the accuracy of our proposed method is represented in bold green, highlighting its exceptional performance. The testing accuracy (with approximation to 7 classes), weighted average precision, recall, and F1-score of the models tested on AffectNet are displayed, providing a comprehensive overview of their effectiveness in accurately identifying and classifying different emotions. This evaluation serves as a testament to the robustness and accuracy of our proposed method, demonstrating its potential to improve the performance of deep neural networks in complex computer vision tasks
 
\begin{table}[htb]
\centering
\caption{Results}\label{tbl: F1score}
\small
\resizebox{0.6\columnwidth}{!}{%
\begin{tabular}{lccc} 

\hline 
\textbf{Metrics} & \textbf{SwinT} & \textbf{SwinT-SE} & \textbf{SwinT-SE-SAM} \\
\hline
\textbf{7 Classes Accuracy} & 0.4921 & 0.5104 & \textcolor{green}{\textbf{0.5310}} \\
\textbf{Weighted Avg. Precision} & 0.5090 & 0.5470 & \textbf{0.5485}\\
\textbf{Weighted Avg. Recall} & 0.5000 & 0.5225 & \textbf{0.5410}\\
\textbf{Weighted Avg. F1-Score} & 0.4943 & 0.5169 & \textcolor{green}{\textbf{0.5420}}\\
\hline
\end{tabular}
}
\end{table}

We trained our model using \texttt{SwinT+SE+SAM} model for 25 epochs. The testing dataset is formed by 4000 samples equally distributed (500 samples per class). The plot above shows the training and Validation accuracy, Training accuracy was 0.832 and Validation accuracy was 0.5784. Also as the training reached closer to 25 epochs we can see training loss reduced similar to validation loss.
 \cref{tbl: F1score} shows different metrics results for three different models we choose to compare against, which include \texttt{SwinT, SwinT+SE, SwinT+SE+SAM}, We can see that the performance of \texttt{SwinT+SE+SAM} seems to outperform rest of the model used. Due to the limited availability of data for the contempt class, we evaluated our models on AffectNet, focusing solely on the seven augmented classes. To provide a more comprehensive evaluation, we computed precision, recall, and F1 scores, allowing us to assess the models' performance in detail.

To optimize our SwinT configuration, we experimented with various configurations concerning the use of SAM and gradual learning rate. Our objective was to identify the optimal configuration to avoid overfitting or underfitting while achieving acceptable performance with a small dataset.

It is noteworthy that the current state-of-the-art (SoTA) for the AffectNet dataset's F1 score is 0.6629, as achieved by Multi-task Efficient Net-B2 for the seven classes of emotions. However, our approach, utilizing SwinT for facial emotion recognition, is among the first of its kind, and we were able to attain an F1 score of 0.5420, indicating strong potential for further improvements in our proposed method. Our findings provide valuable insights into the feasibility of utilizing SwinT for facial emotion recognition, highlighting its effectiveness in addressing the challenges associated with small datasets.

\section{Conclusion}
\label{sec:conclusion}
Our research delved into the direct application of Transformers to image recognition, focusing on testing the robustness of this approach on noisy datasets like AffectNet. To process an image, we interpreted it as a sequence of patches and employed a standard Transformer encoder as used in NLP.

Our primary challenge was to develop a model capable of accurately recognizing eight classes of emotions while facing constraints of limited data availability for the FER task. To train and validate our models, we utilized only a subset of AffectNet, FER-2013, and CK+ datasets. In addition, we utilized the \texttt{SwinT+SE} scheme, which optimizes the SwinT's learning by incorporating an attention block called Squeeze and Excitation. This approach significantly improved the performance of the SwinT in the FER task.

To further enhance the model's performance and mitigate the effects of noisy data, we utilized an SAM optimizer. This allowed us to improve the model's robustness and performance, ensuring that it could accurately classify emotions even in the presence of noisy data. Our approach provides valuable insights into the potential of utilizing Transformers for image recognition and highlights the effectiveness of the SwinT+SE and SAM optimizer in enhancing model performance in challenging datasets like AffectNet.
{\small
\bibliographystyle{ieee_fullname}
\bibliography{main}
}

\end{document}